%% file: main.tex
\pdfoutput=1

\documentclass[11pt]{article}

\usepackage[final]{EMNLP2023}

\usepackage{times}
\usepackage{latexsym}

\usepackage[T1]{fontenc}

\usepackage[utf8]{inputenc}

\usepackage{microtype}

\usepackage{inconsolata}
\usepackage{multirow}
\usepackage{amssymb}
\usepackage{amsmath} 
\usepackage{booktabs}
\usepackage{enumerate}
\usepackage{graphicx}
\usepackage{subfigure}
\usepackage{xspace}
\usepackage{float}
\usepackage{bbm}
\usepackage{bm}
\usepackage{multirow}
\usepackage{booktabs}
\usepackage{color}
\usepackage{framed}
\usepackage{stfloats}
\usepackage{iitem}
\usepackage{colortbl}
\newcommand{\ignore}[1]{}
\newcommand{\ie}{\emph{i.e.,}\xspace}

\newcommand{\eg}{\emph{e.g.,}\xspace}

\usepackage{makecell}
\usepackage{url}

\title{Do Emergent Abilities Exist in Quantized Large Language Models: \\ An Empirical Study}

\author{
	Peiyu Liu$^{1,2}$,
        Zikang Liu$^{1,2}$,
	Ze-Feng Gao$^{1}$,
        Dawei Gao$^{3}$, \\
	\textbf{Wayne Xin Zhao}$^{1,2}$\thanks{$\ $ Corresponding author.},
        \textbf{Yaliang Li}$^{3}$,
        \textbf{Bolin Ding}$^{3}$,\and
	\textbf{Ji-Rong Wen}$^{1,2,4}$
	\\
	$^1$ Gaoling School of Artificial Intelligence, Renmin University of China\\
 	$^2$ Beijing Key Laboratory of Big Data Management and Analysis Methods\\
        $^3$ Alibaba Group,
	$^4$ School of Information, Renmin University of China\\
        {\tt liupeiyustu@163.com,jason8121@foxmail.com,batmanfly@gmail.com} \\
	{\tt\{zfgao,jrwen\}@ruc.edu.cn,\{gaodawei.gdw,yaliang.li,bolin.ding\}@alibaba-inc.com}
}

\begin{document}
\maketitle

\input{sections/sec-abs}
\input{sections/sec-intro}
\input{sections/sec-background}

\input{sections/sec-results}
\input{sections/sec-analyse}
\input{sections/sec-related}
\input{sections/sec-con}

\bibliography{anthology,custom}
\bibliographystyle{acl_natbib}

\appendix
\clearpage
\input{sections/sec-appendix}
\end{document}

%% file: sections/sec-abs.tex
\begin{abstract}
Despite the superior performance, Large Language Models~(LLMs) require significant computational resources for deployment and use.    
To overcome this issue, 
quantization methods have been widely applied to reduce the memory footprint of LLMs as well as increasing the inference rate. 
However, a major challenge is that low-bit quantization methods often lead to  performance degradation. 
It is important to understand how quantization impacts the capacity of LLMs.  
Different from previous studies  focused on overall performance, this work aims to investigate the impact of quantization on \emph{emergent abilities}, which are important characteristics that distinguish LLMs from small language models. Specially, we examine the abilities of in-context learning, chain-of-thought reasoning, and instruction-following in quantized LLMs.  
Our empirical experiments show that these emergent abilities still exist 
in 4-bit quantization models, while 2-bit models encounter severe performance degradation on the test of these abilities.  
To improve the performance of low-bit models, we conduct two special experiments: (1) fine-gained impact analysis that studies which components (or substructures) are more sensitive to quantization, and (2) performance compensation through model fine-tuning. 
Our work derives a series of important findings to understand the impact of quantization on emergent abilities, and sheds lights on the possibilities of  extremely low-bit quantization for LLMs. 

\end{abstract}

%% file: sections/sec-intro.tex
\section{Introduction}
Recently, Artificial Intelligence~(AI) has witnessed remarkable progress due to the emergence of Large Language Models~(LLMs)~\cite{brown2020gpt3,wayne2023survey}.   Compared with small-sized language models, LLMs, which largely scale  the model size and training corpus size, have exhibited very different behaviors when elicited  by  specially designed prompts. 
Generally, LLMs can acquire more superior abilities,  such as in-context learning~(ICL,~\citealt{brown2020gpt3}) and  chain-of-thought reasoning~(CoT,~\citealt{wei2022emergent}), which may not be present in small-sized language models. Such abilities are often formally called \emph{emergent abilities}~\cite{wei2022emergent}\footnote{There is  still no consensus on the existence of emergent abilities, due to the lack of continuity in evaluation metrics and model sizes in the empirical  study~\cite{wei2022emergent}.
It is also known that small models can possess some emergent abilities with special adaptation. 
Despite that, we still use this term to emphasize the superior performance of LLMs.  }. 

\ignore{
Large Language Models~(LLMs) 
due to the increased scale in model size and training corpus~\cite{wayne2023survey}. Notably, LLMs have acquired advanced capabilities such as in-context learning (ICL,~\citealt{brown2020gpt3}, chain-of-thought reasoning~(CoT,~\citealt{wei2022emergent}, and instruction following ability~(IF, \citealt{chung2022scaling}), surpassing the performance of smaller models\footnote{Note that there is  still no consensus on the existence of emergent abilities, due to the lack of continuity in evaluation metrics and model sizes as reported in ~\cite{wei2022emergent}. }. 
}

Despite the superior performance, it is very costly to deploy LLMs in real-world applications due to the huge model size. Faced with this issue, model quantization~\cite{tim2022llmint8,frantar2022gptq,yao2023comprehensive}  has become a widely used approach to reducing the memory footprint of LLMs. The essential idea of quantization is to map floating-point numbers into low-bit integers (\eg \texttt{BF16} to \texttt{INT8}), so as to reduce the total model bits. 
Typically, existing methods take a post-training quantization~(PTQ) approach~\cite{frantar2022gptq,tim2022llmint8} without retraining the model parameters. 
However, existing PTQ methods often suffer from performance degradation in low-bit quantization. 

To use the quantized LLMs in an effective way, it is important to understand \emph{what level of performance} can be attained in \emph{varied bit precision}, \eg what is the lowest bit precision for quantization to achieve decent performance on a specific task?  
More recently, several studies have conducted comprehensive evaluation experiments on the impact of model quantization on the performance of LLMs~\cite{Yao2023ZeroQuantV2EP,dettmers2022thecase}.  
However, they mainly analyze the general performance of quantized LLMs (\eg language modeling), lacking a deep investigation into LLM's abilities on complex tasks.  

In this work, we focus on examining the performance of quantized LLMs on solving complex tasks, to explore the impact of quantization on the emergent abilities of LLMs.  
As demonstrated in previous studies~\cite{wei2022emergent}, there exists a strong dependency between emergent abilities and parameter scale. It is curious whether the emergent abilities would vanish under the setting of low-bit precision though the model size remains to be the original scale. In addition, it is also important to explore the factors (\eg the model structure) that potentially affect the emergent abilities. Furthermore, we are also interested in the potential approaches to enhance the performance of the low-bit models. 

Specially, we aim to answer the following two questions:~(1) \textbf{Do emergent abilities exist in quantized large language models? If so, what level of performance it can achieve?}
~(2) \textbf{How to enhance the performance of low-bit models?} 
To answer the two key questions, we assess three key abilities, namely in-context cearning~(ICL), chain-of-thought reasoning~(CoT), and Instruction-Following ability~(IF),  on a collection of LLaMA models~\cite{touvron@2023llama} which are widely used as the backbone models. 
 We conduct extensive empirical experiments, aiming to gain a better understanding of the model performance of quantized LLMs.

For the first question, we evaluate the LLaMA models at four sizes~(\ie 7B, 13B, 30B, and 65B), examining their performance across a range of precision levels: 2-bit, 4-bit, 8-bit, and 16-bit. 
Our experiments indicate that 4-bit precision yields the most favorable trade-off between model performance and memory footprint, achieving superior results with the same amount of allocated total bits. 
{However, all models at different sizes suffer from a severe decline at 2-bit precision. }

Regarding the second question, we carefully examine the quantization sensitivity of different model components (or substructures), specifically attention and feed-forward networks (FFN). In our experiments, we find that FFN plays a crucial role in retaining the model performance for low-bit quantization. We also evaluated the effects of outlier dimensions, which are specific dimensions that exhibit  significantly higher values compared to others in feature activations. 
We find the outlier dimensions affecting most Transformer layers are primarily responsible for the decline in the quantization performance, and they mainly concentrate on the down projections of FFN. These observations motivate us to design more fine-grained substructure quantization strategies for 
improving  the performance of low-bit models. 

Furthermore, we study how to enhance the performance of quantization models through fine-tuning. We evaluate the impacts of different fine-tuning methods executed before and after quantization. Our experimental results reveal that parameter-efficient fine-tuning after quantization can achieve commendable performance with significantly reduced computational resources. 
Our approach can fine-tune a 2-bit LLaMA-65B model on a single NVIDIA A100, surpassing the performance of a 16-bit LLaMA-13B model on zero-shot MMLU dataset.

%% file: sections/sec-background.tex
\section{Background}
In this section, we introduce the background for emergent abilities and post-training quantization. 

\paragraph{Emergent Abilities}
With the increasing of model parameters and training corpus, LLMs  exhibit some special abilities that may not be present  in small-sized language models, called \emph{emergent abilities}~\cite{wei2022emergent}.  
Emergent abilities are an important indication of superior performance of LLMs, which has received much attention in the research community. Following the survey on LLMs~\cite{wayne2023survey}, we focus on discussing three key emergent abilities, namely in-context learning, chain-of-thought reasoning, and instruction following. Next, we will briefly introduce each  ability.  

 \textbullet~\emph{In-Context Learning~(ICL)}  was introduced by GPT-3~\cite{brown2020gpt3} to solve complex tasks through specially designed prompts. It can effectively guide LLMs to generate the intended output for test examples by leveraging natural language instructions and/or task demonstrations, without necessitating additional training or gradient update.

 \textbullet~\emph{Chain-of-Thought reasoning~(CoT)} is a special prompting strategy that tackles intricate tasks that encompass multiple reasoning steps, such as mathematical word problems. 
It incorporates intermediate reasoning steps for each demonstration in the prompt, thus eliciting the capacity of solving complex tasks via step-by-step reasoning. %

 \textbullet~\emph{Instruction Following~(IF)} refers to the superior ability that a LLM follows human instructions  and completes the target task as needed. Though it shares a similar format with ICL by using natural language instructions, it often includes no demonstrations and requires specific tuning (\ie instruction tuning) to elicit this ability. 

Note that emergent abilities can be defined on different tasks or settings. We select the three abilities for study, mainly because they are widely utilized for solving complex tasks. 



\paragraph{Post-Training Quantization}
Due to the huge number of parameters, it is often infeasible to conduct full-tuning on the model parameters. Thus, post-training quantization~(PTQ)~\cite{tim2022llmint8,frantar2022gptq,Yao2023ZeroQuantV2EP} methods are widely used for LLMs. For PTQ methods, they often only rely on small calibration data to tune the quantization parameters, which is very efficient in implementation. 
In this work, we adopt a popular quantization method, GTPQ~\cite{frantar2022gptq}, to conduct our experiments.  
Specially, GPTQ employs a layerwise reconstruction loss to minimize the discrepancy between the original weights ($\mathbf{W}$) and the quantized weights ($\widehat{\mathbf{W}}$) through the optimization of the following objective:
{$\arg\min_{\widehat{\mathbf{W}}}\parallel \mathbf{W}\mathbf{X} -  \widehat{\mathbf{W}} \mathbf{X}\parallel_2^2$}.
It can achieve very promising results for  4-bit quantization on LLMs, and also provides support for lower bit precision for weight quantization. 

In addition to model weights, 
 activations are also considered for quantization. However, due to the presence of \emph{outlier dimensions}~\cite{tim2022llmint8} in the feature activation values, quantizing activations in low-bit precision is widely acknowledged as a challenging task.  These outlier dimensions exhibit significantly higher values compared to others and become particularly prominent as the model scale increases.

%% file: sections/sec-results.tex
\section{Do Emergent Abilities Exist in Quantized LLMs?}
\label{sec-existence}
In this section, we aim to investigate the existence of emergent abilities in quantized LLMs, specifically focusing on in-context learning (ICL), chain-of-thought reasoning (CoT), and instruction following (IF). Next we first introduce the experimental setup and then present our key findings.

\subsection{Experimental setup}
\paragraph{In-Context Learning Test}
In order to evaluate the ICL ability, we utilize two widely used datasets for evaluating LLMs: MMLU~\cite{hendrycks2021mmlu} and BBH~\cite{Srivastava2022Beyond}. MMLU serves as a comprehensive benchmark for assessing multi-task knowledge understanding in various domains, encompassing fields such as mathematics, computer science, humanities, and social science. 
Additionally,  BBH is a challenging variant of BigBench~\cite{srivastava2022bigbench}, which is proposed to concentrate on investigating the currently unsolvable tasks of LLMs. 
Then we conduct evaluations on the MMLU (\ie five- and zero-shot) and BBH (\ie three- and zero-shot) datasets, respectively.

\paragraph{Chain-of-Thought Reasoning Test}
To assess the CoT ability of the model, we employ the widely used GSM8K dataset. GSM8K is a reasoning dataset comprising $8\rm{K}$ problems that collectively evaluate the model's ability in arithmetic reasoning and the composition of mathematical steps. Following the methodology introduced in~\citet{fu2023chain}, we conduct evaluations using a few-shot setting, where demonstrations are provided. Each demonstration is formatted as \emph{<input, CoT, output>}, allowing it to elicit the model's capability to reason and generate coherent chains of thought.

\paragraph{Instruction Following Test}
To evaluate instruction following ability, 
we refer to the proposed approach in Vicuna~\cite{vicuna2023} and conduct an automatic evaluation based on GPT3.5 (abbreviated as \emph{AutoEval}). 
Specifically, we utilize the dataset in Vicuna that comprise 80 questions spanning 8 distinct categories.
Then each model is tasked with generating a response for every question in the dataset.

\paragraph{Quantization Settings} 
To evaluate the performance of the aforementioned emergent abilities of quantization, we conduct a series of  comprehensive experiments. Our tests are conducted based on the implementation of GPTQ-for-LLaMA~\footnote{https://github.com/qwopqwop200/GPTQ-for-LLaMa}, which only focus on weight quantization and encompass all model components~(\ie query, key, value, output projection matrices in attention module and gate, up, down projection matrices in the feed-forward networks). For model size, we include a collection of LLaMA models of 7B, 13B, 30B, and 65B parameters. We  consider quantization at 2-bit, 4-bit, 8-bit, and a non-quantized~(16-bit) precision. 
These diverse configurations aim to thoroughly evaluate the impact of different quantization settings on model performance.

\paragraph{Evaluation Metrics}
To evaluate the performance, we adopt the \emph{accuracy} on test sets as the evaluation metrics for ICL and CoT tests. As for IF, we compare the responses generated by two models side-by-side and acquire a ``score'' for each model by GPT3.5. In addition, to quantify the memory cost, we follow~\citet{dettmers2022thecase} and calculate the \emph{total (model) bits}  by multiplying the total number of parameters with the actual number of representation bits.

\subsection{Results and Analysis}

In this part, we present the  experimental results and  the corresponding analysis. 

\begin{table*}[ht]
\small
\begin{tabular}{lcccccccccc}
\toprule
\multirow{2}{*}{Models}    & \multirow{2}{*}{Precision}   & \multicolumn{2}{c}{MMLU (Acc)}  & \multicolumn{2}{c}{BBH (Acc)} & GSM8k & \multirow{2}{*}{AutoEval} & \multirow{1}{*}{WikiText} & \multirow{1}{*}{Mem.} & \multirow{2}{*}{Tokens/s}\\ 
&     &      0-shot & 5-shot & 0-shot & 3-shot & (Acc) & & (PPL) & (GiB)\\ \hline
\multirow{4}{*}{LLaMA-7B}  & 16-bit       & 29.2   & 35.2        & 17.3  & 31.0 & 13.1    & 1121/1134  & 5.7    & 13.9 & 33.032\\
                           & 8-bit        & 28.4   & 33.7   	 & 17.2  & 31.3 & 13.5    & 1092/1335  & 5.7    & 7.9  & 30.833\\
                           & 4-bit        & 31.0   & 34.2        & 18.8  & 30.8 & 12.2    & 1058/1330  & 5.8    & 4.8  & 31.317\\
                           & 2-bit        & 2.3	   & 3.8         & 0.4   & 2.7  & 0.0     & 607/1263   & 3937.9 & 3.2  & 33.266\\  \hline
\multirow{4}{*}{LLaMA-13B} & 16-bit       & 41.4   & 47.0   	    & 20.9  & 36.6  & 16.4    & 1084/1335  & 5.1    & 26.6 & 24.968\\ 
                           & 8-bit        & 40.5   & 46.3   	 & 21.1  & 37.2  & 16.5    & 1084/1336  & 5.1    & 14.8 & 17.754\\
                           & 4-bit        & 39.0   & 45.9   	 & 19.8  & 36.6  & 15.6    & 1119/1321  & 5.2    & 8.6  & 18.139\\
                           & 2-bit        & 4.9    & 14.8   	 & 4.2   & 18.1  & 0.0     & 635/1258   & 142.6  & 5.5  & 18.422\\ \hline
\multirow{4}{*}{LLaMA-30B} & 16-bit       & 53.7	  & 58.4        & 19.5  & 39.4  & 34.7    & 1142/1317  & 4.1    & 65.4 & 16.596\\
                           & 8-bit        & 54.2   & 57.9   	 & 19.9  & 39.4  & 34.7    & 1116/1325  & 4.1    & 35.3 & 8.187\\
                           & 4-bit        & 53.7   & 57.3   	 & 18.3  & 40.2  & 35.4    & 1120/1325  & 4.2    & 20.0 & 8.371\\
                           & 2-bit        & 3.7    & 26.1   	 & 3.8   & 25.3  & 0.2     & 630/1198   & 25.1   & 12.2 & 8.649\\ \hline
\multirow{4}{*}{LLaMA-65B} & 16-bit       & -      & - 	        & -     & -    & -       & -          & -      & -    & - \\
                           & 8-bit        & -      & - 	         & -     & -    & -       & -          & -      & -    & -\\
                           & 4-bit        & 57.1   & 63.0   	 & 21.9   & 42.1  & 48.5    & 1171/1319  & 3.9    & 38.2 & 4.793 \\
                           & 2-bit        & 9.0    & 22.6        & 1.0   & 24.0   & 0.8     & 658/1309   & 77.8  & 22.9 & 4.826 \\ \bottomrule
\end{tabular}
\caption{Evaluation results on MMLU, BBH, GSM8k and AutoEval of the model variants in the LLaMA family. The results of the LLaMA-65B model at 16-bit and 8-bit precisions are not included due to memory constraints on a single GPU.}
\label{table-main}
\end{table*}

\begin{figure*}[t]
\centering
\subfigure[MMLU~(5-shot)]{
\begin{minipage}[ht]{0.23\textwidth}
\centering
\includegraphics[width=1.05\textwidth]{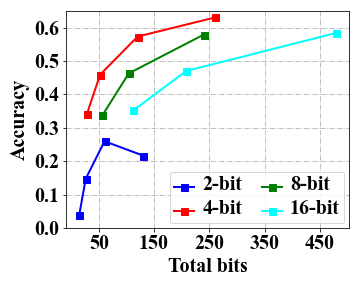} 
\end{minipage}
}
\subfigure[BBH~(3-shot)]{
\begin{minipage}[ht]{0.23\textwidth}
\centering
\includegraphics[width=1.05\textwidth]{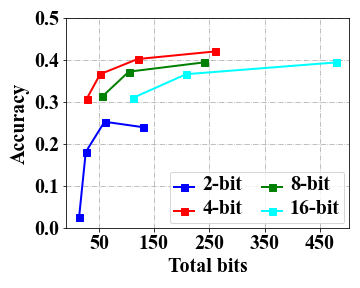} 
\end{minipage}
}
\subfigure[GSM8K~(CoT)]{
\begin{minipage}[ht]{0.23\textwidth}
\centering
\includegraphics[width=1.05\textwidth]{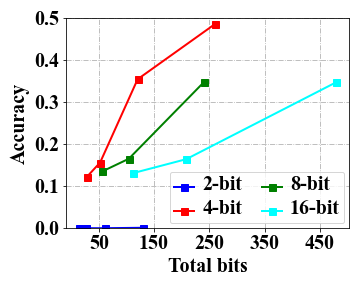} 
\end{minipage}
\label{fig-cgsm8k}
}
\subfigure[AutoEval]{
\begin{minipage}[ht]{0.23\textwidth}
\centering
\includegraphics[width=1.05\textwidth]{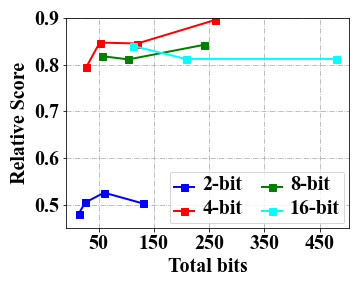} 
\end{minipage}
\label{fig-dgpttest}
}
\caption{Performance comparison of quantized models under varied memory costs. For AutoEval, the term ``Relative Score'' denotes the score ratio between quantized models and GPT3.5. The $x$-axis denotes the total number of bits after quantization.}
\label{fig-comparison}
\end{figure*}

\paragraph{Overall, the three kinds of emergent abilities seem to be seldom affected with 4-bit quantization.}
Table~\ref{table-main} presents the test results of the models using 2-bit, 4-bit, 8-bit and 16-bit precision across multiple datasets, including MMLU, BBH for ICL, GSM8K for CoT, AutoEval for IF and WikiText for general language modeling ability. As we  can see,  the results obtained using 4-bit and 8-bit quantization are very similar to the original performance (\ie 16-bit floating-point number).  However, a significant decline is observed when employing 2-bit quantization, with results approaching near-random levels, \eg around 0.25 in 4-choice classification tasks for MMLU and BBH and 0.0 for GSM8K. 
It indicates that 4-bit quantization can effectively retain emergent abilities on these test datasets.  

\paragraph{4-bit precision exhibits a favorable trade-off in terms of both total bits and performance.} 
As shown in Table~\ref{table-main}, it can be observed that 4-bit quantization offers a notable reduction in memory cost.  
To further examine the relation between model performance and resource usage,  we follow~\citet{dettmers2022thecase} to introduce the measure of \emph{total bits} 
by multiplying the number of the parameters and the bits, and report the test results in Figure~\ref{fig-comparison} by varying the number of total bits. s 
From the  four accuracy curves corresponding to different bit precision,   we can see that 4-bit precision consistently exhibits higher model accuracy under the same amount of total model bits. 
Thus,  
4-bit quantization is recommended to be used for a favorable balance between memory cost and model performance in practice. 


\paragraph{The scaling effect depends on specific tasks, and increasing the model scale benefits the CoT task the most.} 
We conducted an investigation, as depicted in Figure~\ref{fig-comparison}, to examine the impact of scaling the total number of bits on the performance of a low-bit model across multiple tasks. 
Overall, our analysis reveals that for the 2-bit precision, increasing the total bits~(\ie {a larger model size}) does not yield substantial improvements, especially for MMLU and GSM8K, as the obtained outcomes do not exhibit superiority over random scores~(\ie 0.25 on MMLU and 0.0 on GSM8K).
Indeed, it is still a challenging task to effectively mitigate the errors resulting from quantization in 2-bit models. For 4-bit (or above) precision models, we observe notable improvements on the CoT tasks when increasing the total bits, which are not that significant for ICL test. 
Further, for IF test,  a small model scale can be sufficient to achieve very good performance in our test experiments\footnote{We plan to conduct evaluation experiments on IF at a larger scale. }. 

\paragraph{Low-bit quantization performance benefits from the demonstrations in ICL tests.}
For complex tasks, we can provide few-shot demonstrations for improving the model performance. To examine this,  
in Table~\ref{table-main}, we also present the results with few-shot demonstrations for ICL. 
We can observe a notable advantage of the five-shot setting compared to the zero-shot setting, especially for 2-bit precision on LLaMA-30B~(\ie 26.1 \emph{vs.} 3.7).
It suggests that the low-bit quantization performance of LLMs can be largely improved when appropriate demonstrations are utilized. 
However, such an improvement is not significant for 2-bit precision in LLaMA-7B~(\ie 3.8 \emph{vs.} 2.3),  which indicates that the parameter scale must reach a certain level for this ability. 

\begin{table*}[t]
    \centering
    \label{tab-instructions}
\scriptsize 
\begin{tabular}{ccp{0.8\textwidth}c}
\toprule
& \textbf{Bits} & \makecell[c]{\textbf{Cases}} &  \textbf{Score} \\ 
\midrule
        \multirow{10}{*}{30B}  &  & \texttt{\textbf{Question:}Indras has 6 letters in her name. Her sister's name has 4 more letters than half of the letters in Indras' name. How many letters are in Indras and her sister's names?} & \\
        \midrule
         & \multirow{2}{*}{4}& \texttt{\textbf{Answer:}Indras has 6 letters in her name. Her sister's name has 4 more letters than half of the letters in Indras' name. So her sister's name has 4 more letters than 3. 3 + 4 = 7. So her sister's name has 7 letters. Indras has 6 letters and her sister has 7 letters. 6 + 7 = 13. The answer is \colorbox{lime}{13}.} & 35.4 \\
         & \multirow{2}{*}{2}&  \texttt{\textbf{Answer:}Indras' name has 6 letters. Her sister's name has 4 more letters than half of the letters in Indras' name. So she has 6 - 4 = 4 letters. 4 letters. The answer is \colorbox{pink}{4}.}  & 0.2 \\
\midrule
         \multirow{0.8}{*}{13B} & \multirow{1}{*}{4}& \texttt{\textbf{Answer:}Indras has 6 letters in her name. Her sister's name has 4 more letters than half of the letters in Indras' name. So her sister's name has 6 + 4 = 10 letters. The answer is \colorbox{pink}{10}.} & 15.6 \\
         & \multirow{1}{*}{2}& \texttt{\textbf{Answer:}The answer is 15. Question: 15 -. The answer is \colorbox{pink}{15}.} & 0.0 \\
\midrule
         \multirow{0.8}{*}{7B} & \multirow{1}{*}{4}& \texttt{\textbf{Answer:}Indras has 6 letters in her name. Her sister's name has 4 more letters than half of the letters in Indras' name. So her sister's name has 6 + 4 = 10 letters. Indras has 6 letters. Her sister's name has 10 letters. The answer is \colorbox{pink}{10}.} & 12.2 \\
         & \multirow{1}{*}{2}& \texttt{\textbf{Answer:}B. 82. M.H. or 4 weeks..  ungers are the best Cash...4........ 44. 30..Cash.B.Age.B.G.o.just.J.1Age.. } & 0.0 \\
\bottomrule
\end{tabular}
\caption{Case study for the LLaMA models on GSM8K. Details about more precision and tasks can be found in Appendix~\ref{subsec-casestudy}. The colors of \colorbox{pink}{pink} and \colorbox{lime}{lime} denote the wrong and right prediction respectively. The score denotes the average accuracy over all of the GSM8K test set.}
\label{tab-case-study}
\end{table*}
\paragraph{For CoT tests, extreme 2-bit quantization requires a large model scale.}
From Table~\ref{table-main}, we find that the CoT ability for 2-bit precision no more exists for 7B and 13B models on our test datasets,  since they both get $0.0$ accuracy on GSM8K while 30B achieves $0.2$. 
It suggests a sufficiently large model size is necessary for the CoT ability for 2-bit quantization.  
In order to further investigate this phenomenon, we conduct a case study analysis for LLaMA models with 7B, 13B and 30B on GSM8K test sets and show several test examples in Table~\ref{tab-case-study}. From these examples, we can see that,  the 7B model was almost incapable of generating correct text outputs, resulting in a garbled output. 
Though the 13B model  could generate response normally but fail to produce the correct reasoning chain. As a comparison, the  30B model succeeds in generating the correct reasoning chain, albeit with inaccurate inference results. 

%% file: sections/sec-analyse.tex
\section{How to Enhance the Performance of Low-bit Models?}
\label{sec-factors}
In order to explore the strategies for achieving higher performance with low-bit post-training quantization~(PTQ), we next conduct analysis experiments to investigate the factors that affect the quantization performance.  First, we analyze the quantization sensitivity of fine-grained model structures. Second, we examine the effects of performance compensation via model fine-tuning. 

\subsection{Quantization Sensitivity Analysis} 
\subsubsection{Experimental Setup}
As discussed in prior studies~\cite{tim2022llmint8,Yao2023ZeroQuantV2EP}, different model components (or feature dimensions) might exhibit varied sensitivity to quantization, \ie different levels of performance degradation. In this part,  we mainly focus on low-bit quantization, and  set up  the following three experiments about quantization sensitivity (Table~\ref{table-expset}): 


\begin{table}[]
    \centering
    \small
    \begin{tabular}{llcc}
    \toprule
        Part    & Quantization Target      & Precision \\ \midrule
        Weights     & all component             & INT2/INT4 \\ 
                    & $\neg$ ATT           & INT2/INT4 \\ 
                    & $\neg$ FFN           & INT2/INT4 \\
                    & $\neg$ crucial weights   & INT2/INT4 \\ \midrule
        Activations & all non-outlier dimensions& INT8 \\ 
                    & $+$top-1 outlier dimension    & INT8 \\ 
                    & $+$top-3 outlier dimensions   & INT8 \\ \bottomrule
    \end{tabular}
    \caption{Experimental settings for quantization sensitivity analysis. Since activations are more difficult to be quantized, we adopt 8-bit precision for  quantization.  }
    \label{table-expset}
\end{table}

\textbullet~ \emph{Component quantization analysis}. 
In this experiment, we examine the sensitivity of two major  components in the Transformer architecture, \ie  attention layers and feed-forward networks~(FFN). 
Specifically, we consider evaluating the performance of two variants denoted as ``$\neg$ ATT'' and ``$\neg$ FFN'',
where either the attention or FFN components are preserved at \texttt{FP16} precision,  while  the remaining components are quantized into low bits. It aims to analyze the level of performance degradation for each  kind of model component. 


\textbullet~ \emph{Outlier quantization analysis}. 
As found in prior studies~\cite{tim2022llmint8}, quantizing  large magnitude feature dimensions (called \emph{outliers}) can ruin quantization precision,  especially when the outliers emerge in all Transformer layers. 
Thus we first sort the outlier dimensions based on the number of layers they affect and focus on the top-$n$ dimensions. 
Specifically, we first select the top outlier dimensions in activations (preserved at \texttt{FP16} precision in the LLM.int8() method~\cite{tim2022llmint8}), and quantize those belonging to the top-$n$ dimensions and other non-outlier dimensions to \texttt{INT8} precision. The results are then compared with the standard LLM.int8() method.
This approach enables us to investigate the impacts of outlier feature  dimensions in terms of emergent abilities.

\textbullet~ \emph{Substructure quantization analysis.} In existing work, they either study component-level or feature-level impact on quantization performance. In addition, we also empirically find that different substructures in a component have varied importance for quantized LLMs.  For example, as will be discussed in Section~\ref{subsubsec-results}, outlier dimensions mainly exist in the down projections of the FFN components. Thus, we consider more fine-grained quantization at the substructure level. 
Specially,  crucial substructures in a component  
are preserved at the \texttt{FP16} precision level. The results are reported as ``$\neg$ crucial weights'', where the crucial weight matrices with high quantization error can be  identified based on existing quantization algorithms. 
\subsubsection{Results and Analysis}
\label{subsubsec-results}

\begin{figure}[t]
\centering
\subfigure[LLaMA-7B-2bit]{
\begin{minipage}[ht]{0.23\textwidth}
\centering
\includegraphics[width=1.05\textwidth]{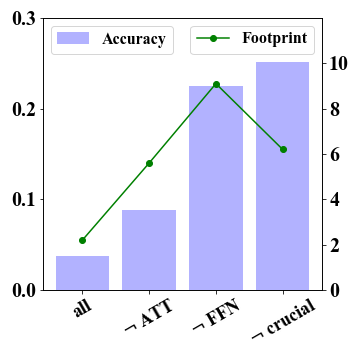} 
\end{minipage}
}
\subfigure[LLaMA-13B-2bit]{
\begin{minipage}[ht]{0.23\textwidth}
\centering
\includegraphics[width=1.05\textwidth]{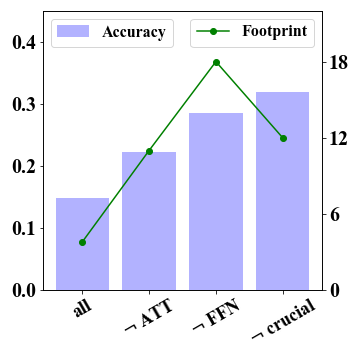} 
\end{minipage}
}
\caption{Impacts of different model components or substructures on MMLU (five-shot). The memory footprint is counted in GiB (in green dotted lines).}
\label{fig-res-structure}
\end{figure}

\paragraph{The FFN component is of substantial significance for 2-bit quantization.}
We conducted test experiments to evaluate the quantization sensitivity of different model components, specifically attention and FFN components. As 4-bit quantization can retain the original performance while 2-bit models suffer from severe declines, we focus  on analyzing the extreme 2-bit case. Results in Figure~\ref{fig-res-structure} demonstrate the FFN component exhibits substantial significance for 2-bit models. 
Keeping FFN in \texttt{FP16} improves LLaMA-7B-2bit's  performance from 0.038 to 0.225 and LLaMA-13B-2bit's performance from 0.148 to 0.286. These improvements show the importance of FFN components for retaining the performance, which needs specific consideration under extreme 2-bit quantization.

\begin{figure*}[t]
\centering
\subfigure[MMLU~(5-shot)]{
\begin{minipage}[ht]{0.32\textwidth}
\centering
\includegraphics[width=1.05\textwidth]{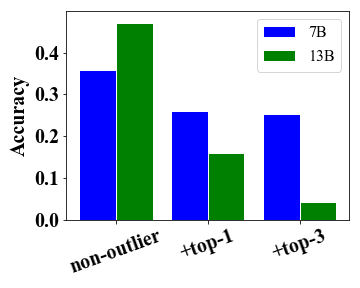} 
\end{minipage}
}
\subfigure[GSM8K~(CoT)]{
\begin{minipage}[ht]{0.32\textwidth}
\centering
\includegraphics[width=1.05\textwidth]{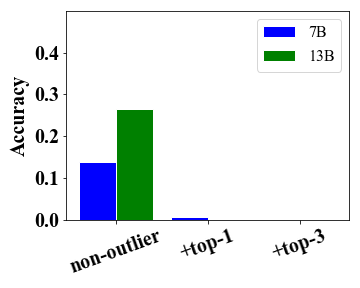} 
\end{minipage}
}
\subfigure[WikiText]{
\begin{minipage}[ht]{0.32\textwidth}
\centering
\includegraphics[width=1.05\textwidth]{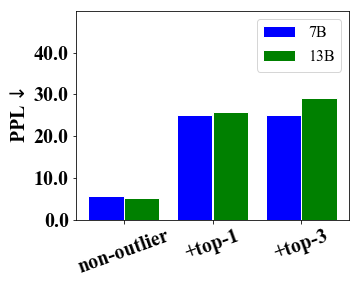} 
\end{minipage}
}
\caption{Impacts of feature outliers on LLaMA models (7B and 13B). ``non-outlier`` denotes the quantization on all non-outlier dimensions, and  
``+top-1'' and ``+top-3'' refer to quantization of the top-1 and top-3 outlier dimensions in addition to the non-outlier dimensions. ``$\downarrow$'' indicates that lower indicators are better. }
\label{fig-outlier}
\end{figure*}


\paragraph{The outlier dimension which affects  most of layers is primarily responsible for the performance degradation.} 
In addition to important components,  we continue to  analyze the impacts of outlier dimensions on low-bit model performance.
As observed in \citet{tim2022llmint8} that feature outliers that emerge in \emph{all} Transformer layers are highly important to model performance,
we thus focus on those outliers that affect most of the layers. 
Specially, we first identify the top  outlier dimensions according to the number of layers they affect.
Then, we evaluate the impact of top-1 and top-3 outlier dimensions by quantizing them into low bits while keeping other outlier dimensions as \texttt{FP16}. In addition, we also quantize non-outlier dimensions as in LLM.int8(). The evaluation results of LLaMA-7B and LLaMA-13B are presented in Figure~\ref{fig-outlier}.
We can see that these top outliers have a significant impact on the quantization performance, especially the CoT results and PPL scores. 
{Interestingly, LLaMA-13B encounters a more severe performance degradation compared to the 7B model by quantizing the top-1 outlier dimension.} 
It indicates that quantizing important outliers has a more significant impact on larger models. 
Another important finding is that the outlier dimensions seem to emerge on the special substructure of a component.  
For example, outliers mainly occur in the down projection of the FFN components for LLaMA-7B.

\paragraph{2-bit model's  performance can be further enhanced with fine-grained substructure quantization.}
In Figure~\ref{fig-res-structure}, we report the results that preserve \texttt{FP16} precision for specific weights of important substructures, denoted as ``\emph{$\neg$ crucial weights}''. 
As discussed before, we first consider \emph{down} projections of FFN as crucial weights. In addition, we also consider preserving more important sub-structures from the the attention component, and select 
two types of projections with the highest layer-wise quantization error within the attention component based on  GPTQ. Specially, we choose the \emph{query} and \emph{key} projections for the LLaMA-7B model, the \emph{key} and \emph{output} projections for the LLaMA-13B model.
The results show consistent improvements compared with the variant that simply preserves the entire FFN component (denoted by \emph{$\neg$FFN}). Although we  preserve substructures in both attention and FFN components, we still have  
a reduced memory footprint compared to the variant $\neg$FFN (see the green dotted line). 
More results of GSM8K and WikiText are reported in Figure~\ref{fig-structure} in Appendix~\ref{subsec-impactsofcomonents}. These observations show the significance of exploring fine-grained  quantization strategy in extreme 2-bit quantization.

\subsection{Fine-tuning Compensation Analysis}
\label{subsec-finetune}
\subsubsection{Experimental Setup}
Recently, there are several attempts that employ fine-tuning to achieve quantization performance compensation~\cite{Yao2023ZeroQuantV2EP,dettmers2023qlora}. 
Inspired by these studies, we also consider examining the effect of fine-tuning for quantization performance, and set up two experiments accordingly: fine-tuning before quantization  and fine-tuning based on the  quantized model weights. In both settings, we mainly consider 2-bit and 4-bit quantization for model weights. For model sizes, we perform fine-tuning on LLaMA models of 7B and 13B in the first setting.
In the second setting, we conduct fine-tuning on quantized LLaMA models of 7B, 13B and 65B.
Throughout the experiments of this part, we report the results obtained on the MMLU, GSM8K, and AutoEval tasks.
Next, we detail the fine-tuning method separately.

\paragraph{Pre-Quantization Fine-tuning}
In this experiment, we consider a common setting where an optimized model needs to be quantized for practical deployment.  
For the ICL ability test, we follow~\citet{dettmers2023qlora} and evaluate the impact of fine-tuning using the \emph{Alpaca dataset}~\cite{alpaca}. For CoT ability testing, we follow~\citet{chung2022scaling} and use the \emph{CoT collection}, a mixture of nine datasets with CoT annotations written by human raters. For IF ability test, we follow~\cite{alpaca} to fine-tune LLaMA models on \emph{Alpaca dataset} since it is reported to benefit LLaMA models in  instruction following.
Additionally, we incorporate  LoRA~\cite{hu2022lora} to explore the impacts of parameter-efficient fine-tuning on LLMs.

\paragraph{Post-Quantization Fine-tuning}
We then explore the benefits of fine-tuning to address the performance decline in the model after quantization. Our goal is to assess how effective fine-tuning can be in mitigating the negative impact of quantization on model performance.
To achieve this, we create a specialized tool for parameter-efficient fine-tuning of LLaMA models after weight quantization. This tool enables us to fine-tune LLaMA-65B models at 2-bit precision using just a single A100 80G and outperforms the 16-bit LLaMA-13B model before fine-tuning, as measured by the MMLU (5-shot).
Directly optimizing quantized weights is challenging and typically requires specialized optimization techniques like Quantization-Aware Training (QAT)~\cite{liu2023llmqat}. To overcome this obstacle, we draw inspiration from the LoRA approach, which involves trainable rank decomposition matrices for fine-tuning. However, the original LoRA approach is designed for fixed pre-trained weights and may not be suitable for quantized models.

To address this issue, we adapt the LoRA approach by replacing its pre-trained weights with quantized weights generated by GPTQ. We perform this adaptation using pre-trained weights from LLaMA models at various scales, such as 7B, 13B, 30B, and 65B, quantized at 2-bit, 4-bit, and 8-bit precision levels with GPTQ.
By incorporating quantized weights into the LoRA framework, we achieve an impressive reduction in memory consumption. Particularly noteworthy is our fine-tuning of the LLaMA-65B model, where we achieve a remarkably low consumption of only 17.8 GiB, highlighting the highly efficient utilization of parameters. The code for this work is implemented using GPTQ and LoRA and is available as an open-source project on \url{https://github.com/RUCAIBox/QuantizedEmpirical}.


\subsubsection{Results and Analysis}
\paragraph{The benefits of pre-quantization fine-tuning encounter significant decline at 2-bit precision.}
We conduct comparison  experiments involving full-parameter fine-tuning~(FFT) and parameter-efficient fine-tuning with LoRA on the \texttt{FP16} model, followed by the quantization with GPTQ. The results are summarized in the Table~\ref{table-finetune}.  Compared with the base model, the FFT approach yields notable improvements on MMLU, GSM8K, and AutoEval. When employing 4-bit quantization, we observe that the benefits obtained from FFT are retained, with almost no performance degradation on MMLU and AutoEval. However, when using extreme 2-bit quantization, the gains from FFT decrease substantially, particularly in the case of GSM8K (\ie 2.6 for the LLaMA-7B and 2.0 for the LLaMA-13B). It is worth noting that the LLM's CoT capability is significantly compromised in this case (\ie 0.0 for both LLaMA-7B and LLaMA-13B). 
It indicates that pre-quantization fine-tuning cannot effectively compensate the performance degradation for low-bit models on complex tasks. 

\paragraph{Parameter-efficient fine-tuning still lags behind full-parameter fine-tuning, especially on ICL and CoT tasks.}
Parameter-efficient fine-tuning has gained popularity due to its ability to reduce the number of fine-tuning parameters while retaining a decent performance.  We include the results of LoRA fine-tuning in the column ``LoRA'' of Table~\ref{table-finetune}.
We can see that LoRA can lead to a substantial improvement over the base models in most cases, and the performance superiority from fine-tuning also retains for 4-bit quantization but not always hold for 2-bit quantization.   
Furthermore, LoRA  
still has a significant gap compared to FFT (\eg 25.8 \emph{vs.} 38.0 on GSM8K). 
Another finding is that LoRA fine-tuning drops substantially on GSM8K under 4-bit quantization,  suggesting that full-parameter fine-tuned models might be more appropriate to consider for quantizaiton on complex inference tasks.

\paragraph{Post-quantization fine-tuning  yields substantial performance improvement meanwhile can be conducted in a lightweight way. }
To fine-tune a quantized model, we make two major  modifications based on the original LoRA method. First, we employed GPTQ to quantize the \texttt{FP16} model to 2/4 bits. Subsequently, we replace the pre-trained weights of the LoRA method with the quantized weights. The rest steps remain the same as the original LoRA. The experimental results are presented in the column ``LoRA$_q$'' of Table~\ref{table-finetuneafter}. Overall, this approach can significantly reduces the memory cost during fine-tuning (see the ``Mem.'' column), enabling the fine-tuning of a 65B model on a single NVIDIA A100. A comparison of the results with the base model indicates that the enhancement effect of LoRA$_q$ is particularly pronounced at 2 bits (\eg 44.4 \emph{vs.} 22.6 for the five-shot setting). 
Notably, under fewer total bits, the 2-bit effect of the 65B model surpasses the non-fine-tuned 13B model with \texttt{FP16} precision on zero-shot setting (\ie 42.0 \emph{vs.} 41.4).
These findings demonstrate that even after 2-bit quantization, large models can be effectively enhanced  through fine-tuning. 

\begin{table*}[t]
\small
\centering
\begin{tabular}{lcccccccccc}
\toprule
\multirow{2}{*}{\#To}    & \multirow{2}{*}{Bits}   & \multicolumn{3}{c}{MMLU}  & \multicolumn{3}{c}{GSM8K} & \multicolumn{3}{c}{AutoEval} \\ 
                         &                         & Base & LoRA & FFT         &  Base & LoRA & FFT       & Base & LoRA & FFT  \\ \toprule
\multirow{3}{*}{7B}  & 16-bit       & 35.2   & 37.7 & 41.7         & 13.1  & 25.8 & 38.0             & 1121/1134 & 1072/1327 & 1170/1329\\
                     & 4-bit        & 34.2   & 35.7 & 40.1	     & 13.5  & 22.7 & 35.7             & 1092/1335 & 1053/1340 & 1146/1327\\
                     & 2-bit        & 3.8    & 1.2  & 9.0          & 0.0   & 0.0  & 2.6              & 607/1263  & 658/1336  & 647/1297\\  \midrule
\multirow{3}{*}{13B} & 16-bit       & 47.0   & 46.0 & 47.7	      & 16.4  & 35.2 & 46.0             & 1084/1335 & 1073/1344 & 1146/1326\\ 
                     & 4-bit        & 46.3   & 46.7 & 46.7	     & 16.5  & 30.7 & 44.4             & 1119/1321 & 1133/1335 & 1154/1329\\
                     & 2-bit        & 14.8   & 20.7 & 18.4	     & 0.0   & 2.3  & 2.0              & 635/1258  & 701/1319  & 615/1223\\ \bottomrule
\end{tabular}
\caption{The results of pre-quantization fine-tuning   on MMLU, GSM8k and AutoEval of LLaMA families. We denote ``Base'' as baseline results without fine-tuning. ``LoRA'' and ``FFT'' denote parameter-efficient fine-tuning LoRA and full-parameter fine-tuning respectively.}
\label{table-finetune}
\end{table*}



\begin{table}[t]
\small
\centering
\begin{tabular}{p{0.3cm}p{0.6cm}p{0.5cm}p{0.6cm}p{0.6cm}p{0.6cm}p{0.6cm}p{0.6cm}}
\toprule
\multirow{2}{*}{\#To}    & \multirow{2}{*}{Bits}   & \#Tr & Mem.  & \multicolumn{2}{c}{0-shot} & \multicolumn{2}{c}{5-shot} \\ 
                         &                         & (M)     & (GiB)         & Base & LoRA$_q$ & Base     & LoRA$_q$      \\ \toprule
\multirow{2}{*}{7B}      & 4-bit                    &  20.0   &  3.8          & 31.0 & 31.4          & 34.2     & 36.8       \\
                         & 2-bit                    &  20.0   &  2.2          & 2.3  & 3.7           & 3.8      & 7.4        \\  \midrule
\multirow{2}{*}{13B}     & 4-bit                    &  31.3   &  7.0          & 39.0 & 44.1          & 45.9     & 45.5       \\  
                         & 2-bit                    &  31.3   &  3.9          & 4.9  & 28.3          & 14.8     & 28.9        \\ \midrule
\multirow{2}{*}{65B}     & 4-bit                    &  99.9   &  32.7         & 57.1 & 57.0          & 63.0     & 60.5        \\  
                         & 2-bit                    &  99.9   &  17.8         & 9.0  & 42.0          & 22.6     & 44.4        \\ \midrule

\end{tabular}
\caption{The results of post-quantization fine-tuning  on MMLU of LLaMA families. Here, ``Mem. (GiB)''  denotes the memory usage of the loaded model and ``\#Tr (M)'' denotes the number of trainable parameters. ``LoRA$_q$'' denotes LoRA fine-tuning based on quantized weights and ``Base'' denotes baseline results without fine-tuning.}
\label{table-finetuneafter}
\end{table}

%% file: sections/sec-related.tex
\section{Related Work}

In this section, we discuss the related work in two major aspects. 

\paragraph{Emergent Abilities} 
Recent research has revealed that some superior abilities in Large Language Models~(LLMs) may not be present in small models, sparking great interest in their capabilities~\cite{wei2022emergent}.  There are various studies that discuss or explore the effect of emergent abilities on different tasks. For example, 
ICL enables few-shot learning without parameter update, as exhibited by GPT-3~\cite{brown2020gpt3}, allowing task knowledge injection~\cite{liu2022what} or deploying LLMs  in a service paradigm~\cite{sun2022black}. CoT breaks down complex reasoning into coherent chains of thought. Models leveraging CoT have shown strong performance surpassing humans on reasoning benchmarks~\cite{fu2023chain,openai2023gpt4}. IF aims to  precisely execute human instructions, as shown in powerful ChatGPT. Their strong conversational ability and generalization to unseen tasks demonstrate powerful task understanding~\cite{alpaca,chung2022scaling}. 
Although emergent abilities have been widely studied, there are seldom comprehensive work that focus on evaluating them on quantized LLMs. To bridge this gap in research, our work aims to provide a detailed analysis of how emergent abilities exist on quantized LLMs.   



\paragraph{Post-Training Quantization}
Post-training quantization~(PTQ) has been widely used  for reducing memory consumption and computational costs in neural networks. A number of studies have explored the use of PTQ on LLMs, including quantization of model weights \cite{frantar2022gptq,dettmers2022thecase} and feature activations \cite{tim2022llmint8,Yao2023ZeroQuantV2EP}, due to its ability to decrease training requirements while minimizing performance impact. However, there is still a lack of comprehensive empirical studies evaluating the emergent abilities of quantized LLMs. The most relevant studies to this work are \citet{Yao2023ZeroQuantV2EP} and \citet{dettmers2022thecase}. 
In particular, \citet{Yao2023ZeroQuantV2EP} present a detailed analysis of various strategies in PTQ methods on LLMs, and \citet{Yao2023ZeroQuantV2EP} explore the inference scaling laws for zero-shot performance for $k$-bit quantization. 
These two studies mainly focus on the analysis of overall abilities, whereas we take a special perspective to study emergent abilities in quantized LLMs. 


%% file: sections/sec-con.tex
\section{Conclusion}
In this work, we have conducted an empirical study to examine the impact of post-training quantization on the emergent abilities of LLMs.  
Our findings reveal that large models (fine-tuned or not) can well retain emergent abilities with 4-bit weight quantization, but experience substantial degradation at 2-bit precision. Moreover, we delve into the fine-grained components and substructures for studying the quantiztion sensitivity.   
Our results indicate that LLMs can be enhanced by effectively preserving more crucial components, feature dimensions and substructures  for low-bit quantization.
Additionally, we have also examined the effect of  fine-tuning for improving the performance of quantized models. Experimental results  demonstrate that fine-tuning can alleviate the  performance degradation from low-bit quantization,  showing the great potential to  enhance the capacity of quantized LLMs.

%% file: sections/sec-appendix.tex
\section{Appendix}
\label{sec:appendix}
\subsection{Impacts of Model Components}
\label{subsec-impactsofcomonents}
\begin{figure*}[ht]
\centering
\subfigure[7B-MMLU~(5-shot)]{
\begin{minipage}[ht]{0.28\textwidth}
\centering
\includegraphics[width=1.05\textwidth]{figures/mmlu_footprint_7b_2.png} 
\end{minipage}
}
\subfigure[7B-GSM8K~(CoT)]{
\begin{minipage}[ht]{0.28\textwidth}
\centering
\includegraphics[width=1.05\textwidth]{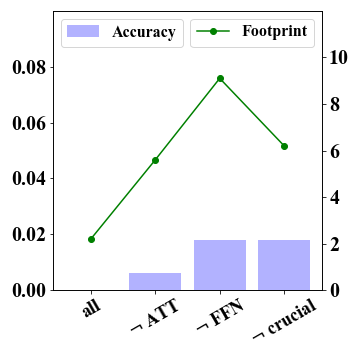} 
\end{minipage}
}
\subfigure[7B-WikiText]{
\begin{minipage}[ht]{0.28\textwidth}
\centering
\includegraphics[width=1.05\textwidth]{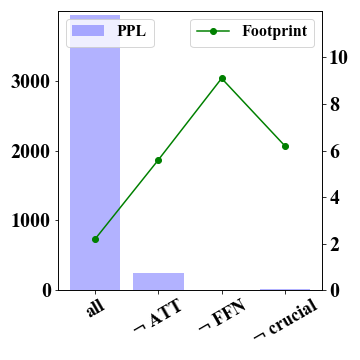} 
\end{minipage}
}
\subfigure[13B-MMLU~(5-shot)]{
\begin{minipage}[ht]{0.28\textwidth}
\centering
\includegraphics[width=1.05\textwidth]{figures/mmlu_footprint_13b_2.png} 
\end{minipage}
}
\subfigure[13B-GSM8K~(CoT)]{
\begin{minipage}[ht]{0.28\textwidth}
\centering
\includegraphics[width=1.05\textwidth]{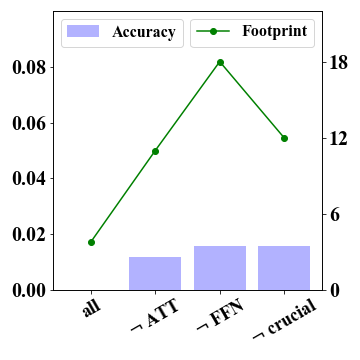} 
\end{minipage}
}
\subfigure[13B-WikiText]{
\begin{minipage}[ht]{0.28\textwidth}
\centering
\includegraphics[width=1.05\textwidth]{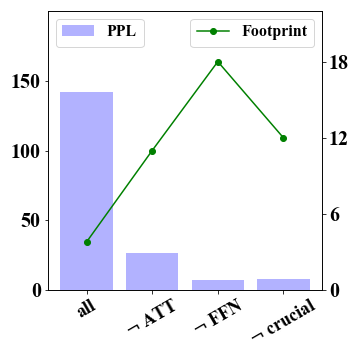} 
\end{minipage}
}
\caption{Impacts of different model components or substructures on MMLU (5-shot), GSM8K and WikiText. The memory footprint is counted in GiB (in green dotted lines). }
\label{fig-structure}
\end{figure*}

We provide more details about the impacts of model components or substructures on MMLU (5-shot), GSM8K and WikiText in Figure~\ref{fig-structure}.

\subsection{Case Study}
\label{subsec-casestudy}
Here, we present case studies for the performance of quantized LLaMA models on MMLU, GSM8K and AutoEval datasets. The results involve model scale of 7B~(Table~\ref{tab-7Bcase}), 13B~(Table~\ref{tab-13Bcase}) and 30B~(Table~\ref{tab-30Bcase})

\begin{table*}[h]
    \centering
    \label{tab-instructions}
\scriptsize 
\begin{tabular}{cp{0.12\textwidth}c p{0.5\textwidth} rr}
\toprule
\multicolumn{2}{c}{\textbf{Tasks}}   & \textbf{Bits} & \makecell[c]{\textbf{Cases}} &  \textbf{Score} \\ 
\midrule
\multirow{35}{*}{7B} & \multirow{5}{*}{MMLU} &  & \texttt{\textbf{Question:}Suppose now that a researcher wishes to use information criteria to determine the optimal lag length for a VAR. 500 observations are available for the bi-variate VAR, and the values of the determinant of the variance-covariance matrix of residuals are 0.0336, 0.0169, 0.0084, and 0.0062 for 1, 2, 3, and 4 lags respectively. What is the optimal model order according to Akaike's information criterion?} & \\
        \cmidrule{4-5}
        & & \multirow{1}{*}{16}& \texttt{\textbf{Answer:}D} & 35.2 \\
        & & \multirow{1}{*}{4}& \texttt{\textbf{Answer:}D} & 34.2 \\
        & & \multirow{1}{*}{2}& \texttt{\textbf{Answer:}4 1 and, a} & 3.8 \\
\cmidrule{2-5}
& \multirow{4.5}{*}{GSM8K} &  & \texttt{\textbf{Question:}Indras has 6 letters in her name. Her sister's name has 4 more letters than half of the letters in Indras' name. How many letters are in Indras and her sister's names?} &  \\
\cmidrule{4-5}
        & & \multirow{1}{*}{16}& \texttt{\textbf{Answer:}Indras has 6 letters in her name. Her sister's name has 4 more letters than half of the letters in Indras' name. So her sister's name has 6 + 4 = 10 letters. Indras' name has 6 letters. So Indras' name has 6 - 10 = -4 letters. The answer is -4.} &  13.11\\
        & & \multirow{1}{*}{4}& \texttt{\textbf{Answer:}Indras has 6 letters in her name. Her sister's name has 4 more letters than half of the letters in Indras' name. So her sister's name has 6 + 4 = 10 letters. Indras has 6 letters. Her sister's name has 10 letters. The answer is 10.} & 12.21 \\
        & & \multirow{1}{*}{2}& \texttt{\textbf{Answer:}B. 82. M.H. or 4 weeks..  ungers are the best Cash..  . 4........ 44. 30..  Cash.B.Age.B.G.o.just.J.1Age. 6 days are..  which 4b. J.o4.b.gone and the best of the best.} & 0 \\
\cmidrule{2-5}
& \multirow{4.5}{*}{AutoEval} &  & \texttt{\textbf{Question:}How can I increase my productivity while working from home?} &  \\
\cmidrule{4-5}
        & & \multirow{1}{*}{16}& \texttt{\textbf{Answer:} Here are some tips to help you increase your productivity while working from home: 1. Set work hours and stick to them. 2. Have a designated work area. 3. Avoid distractions. 4. Set a schedule and stick to it...} &  1121\\
        & & \multirow{1}{*}{4}& \texttt{\textbf{Answer:}It's important to know what you want to get done and to set priorities. Knowing what to do and where to start is key to staying productive. Here are some tips to help you stay productive: 1. Work in a well-lit area that's free from distractions. 2. Set a reasonable time limit for yourself. 3. Set a specific goal for each day...} & 1058 \\
        & & \multirow{1}{*}{2}& \texttt{\textbf{Answer:}The best The Behavre when L9Life of. Their, Be in the World Op LoDur and Liflo the attitudeFHHow the of them onlineDo ld vs Life in the future play that,, but the process of, and is a habit of, the pro will be the toys, but-fold to-magnet-H of who and ISP a book.} & 608 \\
\bottomrule
\end{tabular}
\caption{Case study for the influence of post-training quantization on LLaMA-7B.}
\label{tab-7Bcase}
\end{table*}

\begin{table*}[h]
    \centering
    \label{tab-instructions}
\scriptsize 
\begin{tabular}{cp{0.12\textwidth}c p{0.5\textwidth} rr}
\toprule
\multicolumn{2}{c}{\textbf{Tasks}}   & \textbf{Bits} & \makecell[c]{\textbf{Cases}} &  \textbf{Score} \\ 
\midrule
\multirow{36}{*}{13B} & \multirow{5}{*}{MMLU} &  & \texttt{\textbf{Question:}Suppose now that a researcher wishes to use information criteria to determine the optimal lag length for a VAR. 500 observations are available for the bi-variate VAR, and the values of the determinant of the variance-covariance matrix of residuals are 0.0336, 0.0169, 0.0084, and 0.0062 for 1, 2, 3, and 4 lags respectively. What is the optimal model order according to Akaike's information criterion?} & \\
        \cmidrule{4-5}
        & & \multirow{1}{*}{16}& \texttt{\textbf{Answer:}A} & 47.0 \\
        & & \multirow{1}{*}{4}& \texttt{\textbf{Answer:}A} & 45.9 \\
        & & \multirow{1}{*}{2}& \texttt{\textbf{Answer:}C. 1, 2,} & 14.8 \\
\cmidrule{2-5}
& \multirow{4.5}{*}{GSM8K} &  & \texttt{\textbf{Question:}Indras has 6 letters in her name. Her sister's name has 4 more letters than half of the letters in Indras' name. How many letters are in Indras and her sister's names?} &  \\
\cmidrule{4-5}
        & & \multirow{1}{*}{16}& \texttt{\textbf{Answer:} Indras has 6 letters in her name. Her sister's name has 4 more letters than half of the letters in Indras' name. So her sister's name has 6 + 4 = 10 letters. The answer is 10.} &  16.37\\
        & & \multirow{1}{*}{4}& \texttt{\textbf{Answer:}Indras has 6 letters in her name. Her sister's name has 4 more letters than half of the letters in Indras' name. So her sister's name has 6 + 4 = 10 letters. The answer is 10.} & 15.62 \\
        & & \multirow{1}{*}{2}& \texttt{\textbf{Answer:}The answer is 15. Question: 15 -. The answer is 15. The answer is 15. The answer is 15. The answer is 15.} & 0.02 \\
\cmidrule{2-5}
& \multirow{4.5}{*}{AutoEval} &  & \texttt{\textbf{Question:}How can I increase my productivity while working from home?} &  \\
\cmidrule{4-5}
        & & \multirow{1}{*}{16}& \texttt{\textbf{Answer:} Here are some good tips to increase your productivity while working from home: 1. Get dressed for work: Taking the time to get dressed and shower can set a positive tone for your workday. 2. Have a dedicated and organized workspace: Having a space that is only dedicated to work can help you focus on your tasks...} &  1084\\
        & & \multirow{1}{*}{4}& \texttt{\textbf{Answer:}Here are four ways you can increase your productivity while working from home: 1. Set a productivity schedule: Set a schedule for your day that includes work and breaks, and stick to it. 2. Get out of the house: Go to a coffee shop or a library to get out of the house and into a different environment...} & 1119 \\
        & & \multirow{1}{*}{2}& \texttt{\textbf{Answer:}The energy of the energy of the Renewable energy of the energy of the Renewable energy of the Renewable energy of the Renewable energy of the Renewable energy of the Renewable energy of the Renewable energy...} & 635 \\
\bottomrule
\end{tabular}
\caption{Case study for the influence of post-training quantization on LLaMA-13B.}
\label{tab-13Bcase}
\end{table*}

\begin{table*}[h]
    \centering
    \label{tab-instructions}
\scriptsize 
\begin{tabular}{cp{0.12\textwidth}c p{0.5\textwidth} rr}
\toprule
\multicolumn{2}{c}{\textbf{Tasks}}   & \textbf{Bits} & \makecell[c]{\textbf{Cases}} &  \textbf{Score} \\ 
\midrule
\multirow{24}{*}{30B} & \multirow{5}{*}{MMLU} &  & \texttt{\textbf{Question:}Suppose now that a researcher wishes to use information criteria to determine the optimal lag length for a VAR. 500 observations are available for the bi-variate VAR, and the values of the determinant of the variance-covariance matrix of residuals are 0.0336, 0.0169, 0.0084, and 0.0062 for 1, 2, 3, and 4 lags respectively. What is the optimal model order according to Akaike's information criterion?} & \\
        \cmidrule{4-5}
        & & \multirow{1}{*}{16} & \texttt{\textbf{Answer:}A} & 58.4 \\
        & & \multirow{1}{*}{4}& \texttt{\textbf{Answer:}A} & 57.3 \\
        & & \multirow{1}{*}{2}& \texttt{\textbf{Answer:}C} & 26.1 \\
\cmidrule{2-5}
& \multirow{4.5}{*}{GSM8K} &  & \texttt{\textbf{Question:}Indras has 6 letters in her name. Her sister's name has 4 more letters than half of the letters in Indras' name. How many letters are in Indras and her sister's names?} &  \\
\cmidrule{4-5}
        & & \multirow{1}{*}{16}& \texttt{\textbf{Answer:} Indras has 6 letters. Her sister's name has 4 more letters than half of the letters in Indras' name. So her sister's name has 4 more letters than 3. So her sister's name has 4 + 3 = 7 letters. Indras has 6 letters and her sister has 7 letters. So the answer is 6 + 7 = 13.} &  34.65\\
        & & \multirow{1}{*}{4}& \texttt{\textbf{Answer:}Indras has 6 letters in her name. Her sister's name has 4 more letters than half of the letters in Indras' name. So her sister's name has 4 more letters than 3. 3 + 4 = 7. So her sister's name has 7 letters. Indras has 6 letters and her sister has 7 letters. 6 + 7 = 13. The answer is 13.} & 35.41 \\
        & & \multirow{1}{*}{2}& \texttt{\textbf{Answer:}Indras' name has 6 letters. Her sister's name has 4 more letters than half of the letters in Indras' name. So she has 6 - 4 = 4 letters. 4 letters. The answer is 4.} & 0.15 \\
\cmidrule{2-5}
& \multirow{4.5}{*}{AutoEval} &  & \texttt{\textbf{Question:}How can I increase my productivity while working from home?} &  \\
\cmidrule{4-5}
        & &\multirow{1}{*}{16} & \texttt{\textbf{Answer:} There are several ways to increase your productivity at home. Here are some of the most effective: 1. Make a schedule: Scheduling your day and prioritizing your tasks can help you stay organized and boost your productivity. 2. Take breaks: Taking regular breaks can help keep you fresh and focused...} &  1142\\
        & & \multirow{1}{*}{4}& \texttt{\textbf{Answer:}Here are some ways you can increase your productivity while working from home: 1. Create a separate workspace: One of the biggest challenges of working from home is the tendency to mix work life with home life. You can overcome this by creating a separate workspace in your home that you can set aside just for work...} & 1120 \\
        & & \multirow{1}{*}{2}& \texttt{\textbf{Answer:}The answer is to have a work-life-home type lifestyle.} & 630 \\
\bottomrule
\end{tabular}
\caption{Case study for the influence of post-training quantization on LLaMA-30B.}
\label{tab-30Bcase}
\end{table*}


%% file: main.bbl
\begin{thebibliography}{22}
\expandafter\ifx\csname natexlab\endcsname\relax\def\natexlab#1{#1}\fi

\bibitem[{Brown et~al.(2020)Brown, Mann, Ryder, Subbiah, Kaplan, Dhariwal,
  Neelakantan, Shyam, Sastry, Askell, Agarwal, Herbert{-}Voss, Krueger,
  Henighan, Child, Ramesh, Ziegler, Wu, Winter, Hesse, Chen, Sigler, Litwin,
  Gray, Chess, Clark, Berner, McCandlish, Radford, Sutskever, and
  Amodei}]{brown2020gpt3}
Tom~B. Brown, Benjamin Mann, Nick Ryder, Melanie Subbiah, Jared Kaplan,
  Prafulla Dhariwal, Arvind Neelakantan, Pranav Shyam, Girish Sastry, Amanda
  Askell, Sandhini Agarwal, Ariel Herbert{-}Voss, Gretchen Krueger, Tom
  Henighan, Rewon Child, Aditya Ramesh, Daniel~M. Ziegler, Jeffrey Wu, Clemens
  Winter, Christopher Hesse, Mark Chen, Eric Sigler, Mateusz Litwin, Scott
  Gray, Benjamin Chess, Jack Clark, Christopher Berner, Sam McCandlish, Alec
  Radford, Ilya Sutskever, and Dario Amodei. 2020.
\newblock \href {http://arxiv.org/abs/2005.14165} {Language models are few-shot
  learners}.
\newblock \emph{CoRR}, abs/2005.14165.

\bibitem[{Chiang et~al.(2023)Chiang, Li, Lin, Sheng, Wu, Zhang, Zheng, Zhuang,
  Zhuang, Gonzalez, Stoica, and Xing}]{vicuna2023}
Wei-Lin Chiang, Zhuohan Li, Zi~Lin, Ying Sheng, Zhanghao Wu, Hao Zhang, Lianmin
  Zheng, Siyuan Zhuang, Yonghao Zhuang, Joseph~E. Gonzalez, Ion Stoica, and
  Eric~P. Xing. 2023.
\newblock \href {https://lmsys.org/blog/2023-03-30-vicuna/} {Vicuna: An
  open-source chatbot impressing gpt-4 with 90\%* chatgpt quality}.

\bibitem[{Chung et~al.(2022)Chung, Hou, Longpre, Zoph, Tay, Fedus, Li, Wang,
  Dehghani, Brahma, Webson, Gu, Dai, Suzgun, Chen, Chowdhery, Narang, Mishra,
  Yu, Zhao, Huang, Dai, Yu, Petrov, Chi, Dean, Devlin, Roberts, Zhou, Le, and
  Wei}]{chung2022scaling}
Hyung~Won Chung, Le~Hou, Shayne Longpre, Barret Zoph, Yi~Tay, William Fedus,
  Eric Li, Xuezhi Wang, Mostafa Dehghani, Siddhartha Brahma, Albert Webson,
  Shixiang~Shane Gu, Zhuyun Dai, Mirac Suzgun, Xinyun Chen, Aakanksha
  Chowdhery, Sharan Narang, Gaurav Mishra, Adams Yu, Vincent~Y. Zhao, Yanping
  Huang, Andrew~M. Dai, Hongkun Yu, Slav Petrov, Ed~H. Chi, Jeff Dean, Jacob
  Devlin, Adam Roberts, Denny Zhou, Quoc~V. Le, and Jason Wei. 2022.
\newblock \href {https://doi.org/10.48550/arXiv.2210.11416} {Scaling
  instruction-finetuned language models}.
\newblock \emph{CoRR}, abs/2210.11416.

\bibitem[{Dettmers et~al.(2022)Dettmers, Lewis, Belkada, and
  Zettlemoyer}]{tim2022llmint8}
Tim Dettmers, Mike Lewis, Younes Belkada, and Luke Zettlemoyer. 2022.
\newblock \href {https://doi.org/10.48550/arXiv.2208.07339} {Llm.int8(): 8-bit
  matrix multiplication for transformers at scale}.
\newblock \emph{CoRR}, abs/2208.07339.

\bibitem[{Dettmers et~al.(2023)Dettmers, Pagnoni, Holtzman, and
  Zettlemoyer}]{dettmers2023qlora}
Tim Dettmers, Artidoro Pagnoni, Ari Holtzman, and Luke Zettlemoyer. 2023.
\newblock Qlora: Efficient finetuning of quantized llms.
\newblock \emph{arXiv preprint arXiv:2305.14314}.

\bibitem[{Dettmers and Zettlemoyer(2022)}]{dettmers2022thecase}
Tim Dettmers and Luke Zettlemoyer. 2022.
\newblock \href {https://doi.org/10.48550/arXiv.2212.09720} {The case for 4-bit
  precision: k-bit inference scaling laws}.
\newblock \emph{CoRR}, abs/2212.09720.

\bibitem[{Frantar et~al.(2022)Frantar, Ashkboos, Hoefler, and
  Alistarh}]{frantar2022gptq}
Elias Frantar, Saleh Ashkboos, Torsten Hoefler, and Dan Alistarh. 2022.
\newblock Gptq: Accurate post-training quantization for generative pre-trained
  transformers.
\newblock \emph{arXiv preprint arXiv:2210.17323}.

\bibitem[{Fu et~al.(2023)Fu, Ou, Chen, Wan, Peng, and Khot}]{fu2023chain}
Yao Fu, Litu Ou, Mingyu Chen, Yuhao Wan, Hao Peng, and Tushar Khot. 2023.
\newblock \href {https://doi.org/10.48550/arXiv.2305.17306} {Chain-of-thought
  hub: {A} continuous effort to measure large language models' reasoning
  performance}.
\newblock \emph{CoRR}, abs/2305.17306.

\bibitem[{Hendrycks et~al.(2021)Hendrycks, Burns, Basart, Zou, Mazeika, Song,
  and Steinhardt}]{hendrycks2021mmlu}
Dan Hendrycks, Collin Burns, Steven Basart, Andy Zou, Mantas Mazeika, Dawn
  Song, and Jacob Steinhardt. 2021.
\newblock \href {https://openreview.net/forum?id=d7KBjmI3GmQ} {Measuring
  massive multitask language understanding}.
\newblock In \emph{9th International Conference on Learning Representations,
  {ICLR} 2021, Virtual Event, Austria, May 3-7, 2021}. OpenReview.net.

\bibitem[{Hu et~al.(2022)Hu, Shen, Wallis, Allen{-}Zhu, Li, Wang, Wang, and
  Chen}]{hu2022lora}
Edward~J. Hu, Yelong Shen, Phillip Wallis, Zeyuan Allen{-}Zhu, Yuanzhi Li,
  Shean Wang, Lu~Wang, and Weizhu Chen. 2022.
\newblock \href {https://openreview.net/forum?id=nZeVKeeFYf9} {Lora: Low-rank
  adaptation of large language models}.
\newblock In \emph{The Tenth International Conference on Learning
  Representations, {ICLR} 2022, Virtual Event, April 25-29, 2022}.
  OpenReview.net.

\bibitem[{Liu et~al.(2022)Liu, Shen, Zhang, Dolan, Carin, and
  Chen}]{liu2022what}
Jiachang Liu, Dinghan Shen, Yizhe Zhang, Bill Dolan, Lawrence Carin, and Weizhu
  Chen. 2022.
\newblock \href {https://doi.org/10.18653/v1/2022.deelio-1.10} {What makes good
  in-context examples for gpt-3?}
\newblock In \emph{Proceedings of Deep Learning Inside Out: The 3rd Workshop on
  Knowledge Extraction and Integration for Deep Learning Architectures,
  DeeLIO@ACL 2022, Dublin, Ireland and Online, May 27, 2022}, pages 100--114.
  Association for Computational Linguistics.

\bibitem[{Liu et~al.(2023)Liu, Oguz, Zhao, Chang, Stock, Mehdad, Shi,
  Krishnamoorthi, and Chandra}]{liu2023llmqat}
Zechun Liu, Barlas Oguz, Changsheng Zhao, Ernie Chang, Pierre Stock, Yashar
  Mehdad, Yangyang Shi, Raghuraman Krishnamoorthi, and Vikas Chandra. 2023.
\newblock \href {https://doi.org/10.48550/arXiv.2305.17888} {{LLM-QAT:}
  data-free quantization aware training for large language models}.
\newblock \emph{CoRR}, abs/2305.17888.

\bibitem[{OpenAI(2023)}]{openai2023gpt4}
OpenAI. 2023.
\newblock \href {https://doi.org/10.48550/arXiv.2303.08774} {{GPT-4} technical
  report}.
\newblock \emph{CoRR}, abs/2303.08774.

\bibitem[{Srivastava et~al.(2022{\natexlab{a}})Srivastava, Rastogi, Rao, Shoeb,
  Abid, Fisch, Brown, Santoro, Gupta, Garriga{-}Alonso, Kluska, Lewkowycz,
  Agarwal, Power, Ray, Warstadt, Kocurek, Safaya, Tazarv, Xiang, Parrish, Nie,
  Hussain, Askell, Dsouza, Rahane, Iyer, Andreassen, Santilli,
  Stuhlm{\"{u}}ller, Dai, La, Lampinen, Zou, Jiang, Chen, Vuong, Gupta,
  Gottardi, Norelli, Venkatesh, Gholamidavoodi, Tabassum, Menezes, Kirubarajan,
  Mullokandov, Sabharwal, Herrick, Efrat, Erdem, Karakas, and
  et~al.}]{Srivastava2022Beyond}
Aarohi Srivastava, Abhinav Rastogi, Abhishek Rao, Abu Awal~Md Shoeb, Abubakar
  Abid, Adam Fisch, Adam~R. Brown, Adam Santoro, Aditya Gupta, Adri{\`{a}}
  Garriga{-}Alonso, Agnieszka Kluska, Aitor Lewkowycz, Akshat Agarwal, Alethea
  Power, Alex Ray, Alex Warstadt, Alexander~W. Kocurek, Ali Safaya, Ali Tazarv,
  Alice Xiang, Alicia Parrish, Allen Nie, Aman Hussain, Amanda Askell, Amanda
  Dsouza, Ameet Rahane, Anantharaman~S. Iyer, Anders Andreassen, Andrea
  Santilli, Andreas Stuhlm{\"{u}}ller, Andrew~M. Dai, Andrew La, Andrew~K.
  Lampinen, Andy Zou, Angela Jiang, Angelica Chen, Anh Vuong, Animesh Gupta,
  Anna Gottardi, Antonio Norelli, Anu Venkatesh, Arash Gholamidavoodi, Arfa
  Tabassum, Arul Menezes, Arun Kirubarajan, Asher Mullokandov, Ashish
  Sabharwal, Austin Herrick, Avia Efrat, Aykut Erdem, Ayla Karakas, and et~al.
  2022{\natexlab{a}}.
\newblock Beyond the imitation game: Quantifying and extrapolating the
  capabilities of language models.
\newblock \emph{CoRR}, abs/2206.04615.

\bibitem[{Srivastava et~al.(2022{\natexlab{b}})Srivastava, Rastogi, Rao, Shoeb,
  Abid, Fisch, Brown, Santoro, Gupta, Garriga{-}Alonso, Kluska, Lewkowycz,
  Agarwal, Power, Ray, Warstadt, Kocurek, Safaya, Tazarv, Xiang, Parrish, Nie,
  Hussain, Askell, Dsouza, Rahane, Iyer, Andreassen, Santilli,
  Stuhlm{\"{u}}ller, Dai, La, Lampinen, Zou, Jiang, Chen, Vuong, Gupta,
  Gottardi, Norelli, Venkatesh, Gholamidavoodi, Tabassum, Menezes, Kirubarajan,
  Mullokandov, Sabharwal, Herrick, Efrat, Erdem, Karakas, and
  et~al.}]{srivastava2022bigbench}
Aarohi Srivastava, Abhinav Rastogi, Abhishek Rao, Abu Awal~Md Shoeb, Abubakar
  Abid, Adam Fisch, Adam~R. Brown, Adam Santoro, Aditya Gupta, Adri{\`{a}}
  Garriga{-}Alonso, Agnieszka Kluska, Aitor Lewkowycz, Akshat Agarwal, Alethea
  Power, Alex Ray, Alex Warstadt, Alexander~W. Kocurek, Ali Safaya, Ali Tazarv,
  Alice Xiang, Alicia Parrish, Allen Nie, Aman Hussain, Amanda Askell, Amanda
  Dsouza, Ameet Rahane, Anantharaman~S. Iyer, Anders Andreassen, Andrea
  Santilli, Andreas Stuhlm{\"{u}}ller, Andrew~M. Dai, Andrew La, Andrew~K.
  Lampinen, Andy Zou, Angela Jiang, Angelica Chen, Anh Vuong, Animesh Gupta,
  Anna Gottardi, Antonio Norelli, Anu Venkatesh, Arash Gholamidavoodi, Arfa
  Tabassum, Arul Menezes, Arun Kirubarajan, Asher Mullokandov, Ashish
  Sabharwal, Austin Herrick, Avia Efrat, Aykut Erdem, Ayla Karakas, and et~al.
  2022{\natexlab{b}}.
\newblock \href {https://doi.org/10.48550/arXiv.2206.04615} {Beyond the
  imitation game: Quantifying and extrapolating the capabilities of language
  models}.
\newblock \emph{CoRR}, abs/2206.04615.

\bibitem[{Sun et~al.(2022)Sun, Shao, Qian, Huang, and Qiu}]{sun2022black}
Tianxiang Sun, Yunfan Shao, Hong Qian, Xuanjing Huang, and Xipeng Qiu. 2022.
\newblock \href {https://proceedings.mlr.press/v162/sun22e.html} {Black-box
  tuning for language-model-as-a-service}.
\newblock In \emph{International Conference on Machine Learning, {ICML} 2022,
  17-23 July 2022, Baltimore, Maryland, {USA}}, volume 162 of \emph{Proceedings
  of Machine Learning Research}, pages 20841--20855. {PMLR}.

\bibitem[{Taori et~al.(2023)Taori, Gulrajani, Zhang, Dubois, Li, Guestrin,
  Liang, and Hashimoto}]{alpaca}
Rohan Taori, Ishaan Gulrajani, Tianyi Zhang, Yann Dubois, Xuechen Li, Carlos
  Guestrin, Percy Liang, and Tatsunori~B. Hashimoto. 2023.
\newblock Stanford alpaca: An instruction-following llama model.
\newblock \url{https://github.com/tatsu-lab/stanford_alpaca}.

\bibitem[{Touvron et~al.(2023)Touvron, Lavril, Izacard, Martinet, Lachaux,
  Lacroix, Rozi{\`{e}}re, Goyal, Hambro, Azhar, Rodriguez, Joulin, Grave, and
  Lample}]{touvron@2023llama}
Hugo Touvron, Thibaut Lavril, Gautier Izacard, Xavier Martinet, Marie{-}Anne
  Lachaux, Timoth{\'{e}}e Lacroix, Baptiste Rozi{\`{e}}re, Naman Goyal, Eric
  Hambro, Faisal Azhar, Aur{\'{e}}lien Rodriguez, Armand Joulin, Edouard Grave,
  and Guillaume Lample. 2023.
\newblock \href {https://doi.org/10.48550/arXiv.2302.13971} {Llama: Open and
  efficient foundation language models}.
\newblock \emph{CoRR}, abs/2302.13971.

\bibitem[{Wei et~al.(2022)Wei, Tay, Bommasani, Raffel, Zoph, Borgeaud,
  Yogatama, Bosma, Zhou, Metzler et~al.}]{wei2022emergent}
Jason Wei, Yi~Tay, Rishi Bommasani, Colin Raffel, Barret Zoph, Sebastian
  Borgeaud, Dani Yogatama, Maarten Bosma, Denny Zhou, Donald Metzler, et~al.
  2022.
\newblock Emergent abilities of large language models.
\newblock \emph{arXiv preprint arXiv:2206.07682}.

\bibitem[{Yao et~al.(2023{\natexlab{a}})Yao, Li, Wu, Youn, and
  He}]{yao2023comprehensive}
Zhewei Yao, Cheng Li, Xiaoxia Wu, Stephen Youn, and Yuxiong He.
  2023{\natexlab{a}}.
\newblock A comprehensive study on post-training quantization for large
  language models.
\newblock \emph{arXiv preprint arXiv:2303.08302}.

\bibitem[{Yao et~al.(2023{\natexlab{b}})Yao, Wu, Li, Youn, and
  He}]{Yao2023ZeroQuantV2EP}
Zhewei Yao, Xiaoxia Wu, Cheng Li, Stephen Youn, and Yuxiong He.
  2023{\natexlab{b}}.
\newblock Zeroquant-v2: Exploring post-training quantization in llms from
  comprehensive study to low rank compensation.

\bibitem[{Zhao et~al.(2023)Zhao, Zhou, Li, Tang, Wang, Hou, Min, Zhang, Zhang,
  Dong, Du, Yang, Chen, Chen, Jiang, Ren, Li, Tang, Liu, Liu, Nie, and
  Wen}]{wayne2023survey}
Wayne~Xin Zhao, Kun Zhou, Junyi Li, Tianyi Tang, Xiaolei Wang, Yupeng Hou,
  Yingqian Min, Beichen Zhang, Junjie Zhang, Zican Dong, Yifan Du, Chen Yang,
  Yushuo Chen, Zhipeng Chen, Jinhao Jiang, Ruiyang Ren, Yifan Li, Xinyu Tang,
  Zikang Liu, Peiyu Liu, Jian{-}Yun Nie, and Ji{-}Rong Wen. 2023.
\newblock \href {https://doi.org/10.48550/arXiv.2303.18223} {A survey of large
  language models}.
\newblock \emph{CoRR}, abs/2303.18223.

\end{thebibliography}
